\documentclass[11pt]{article}
\usepackage[utf8]{inputenc}
\usepackage[margin=1in]{geometry}

\usepackage{microtype}
\usepackage{graphicx}
\usepackage{booktabs}
\usepackage{natbib}
\setcitestyle{open={(},close={)}}

\usepackage[colorlinks=true,citecolor=blue]{hyperref}

\usepackage{epsfig}
\usepackage{graphicx}
\usepackage{amsmath}
\usepackage{amssymb}
\usepackage{booktabs}

\usepackage{multirow}
\usepackage{booktabs,amsfonts,dcolumn}
\usepackage{colortbl}
\definecolor{mygray}{gray}{0.95}
\usepackage{tabulary}
\newcolumntype{x}[1]{>{\centering\arraybackslash}p{#1pt}}
\usepackage{bm}
\usepackage{amsmath}
\usepackage{wrapfig}
\usepackage{caption}
\usepackage{subcaption}

\def\eqref#1{equation~\ref{#1}}

\def\1{\bm{1}}

\DeclareMathAlphabet{\mathsfit}{\encodingdefault}{\sfdefault}{m}{sl}
\SetMathAlphabet{\mathsfit}{bold}{\encodingdefault}{\sfdefault}{bx}{n}

\renewcommand{\paragraph}[1]{\vspace{1.25mm}\noindent\textbf{#1}}
\newcommand{\eg}{\textit{e}.\textit{g}. }

\begin{document}

\title{\bf R$^2$-MLP: Round-Roll MLP Architecture for\\ Multi-View 3D Object Recognition}

\author{\vspace{0.5in}\\\bf Shuo Chen, Tan Yu, Ping Li\\\\
Cognitive Computing Lab\\
Baidu Research \\
10900 NE 8th St. Bellevue, WA 98004, USA\\\\
\{shanshuo1992, tan.yu1503, pingli98\}@gmail.com
}

\date{}
\maketitle

\begin{abstract}\vspace{0.3in}

\noindent Recently, vision architectures based exclusively on multi-layer perceptrons (MLPs) have gained much attention in the computer vision community. MLP-like models achieve competitive performance on a single 2D image classification with less inductive bias without hand-crafted convolution layers. In this work, we explore the effectiveness of MLP-based architecture for the view-based 3D object recognition task. We present an MLP-based architecture termed as Round-Roll MLP (R$^2$-MLP). It extends the spatial-shift MLP backbone by considering the communications between patches from different views. R$^2$-MLP rolls part of the channels along the view dimension and promotes information exchange between neighboring views. We benchmark MLP results on ModelNet10 and ModelNet40 datasets with ablations in various aspects. The experimental results show that, with a conceptually simple structure, our R$^2$-MLP achieves competitive performance compared with existing state-of-the-art methods.\vspace{-0.1in}
\end{abstract}

\newpage

\section{Introduction}
With the increase and maturity of 3D acquisition technology, 3D object recognition has become one of the most popular research directions in object recognition. In the past decade, we have witnessed the great success of convolutional neural networks (CNNs) in image understanding. CNNs have been widely used in 3D object classification and achieved excellent performance. 3D object recognition methods based on CNNs can be divided into three research directions~\citep{yang2019learning} by the input mode, namely, voxel-based methods~\citep{wu20153d,qi2016volumetric}, point-based methods~\citep{qi2017pointnet,qi2017pointnet++}, and view-based methods~\citep{wang2017dominant,feng2018gvcnn,kanezaki2018rotationnet,han2019seqview2seqlabels,han20193d2seqvies,wei2020view}. Among them, view-based methods render each 3D objects to 2D images from different viewpoints and model each view through the CNN originally devised for modelling 2D images. Since view-based methods can fine-tune from the model pre-trained on large-scale 2D image recognition datasets such as ImageNet, they normally perform better than their voxel-based and point-based counterparts training from scratch.

Many efforts have been devoted to improving the effectiveness of view-based methods for 3D object recognition in the past few years. MVCNN~\citep{su2015multi} is one of the earliest attempts to adapt CNN to model the projected views of a 3D object. It directly max-pools the view features from a 2D CNN into a global 3D object representation for recognition. Following MVCNN~\citep{su2015multi}, these works~\citep{feng2018gvcnn,han20193d2seqvies,han2019seqview2seqlabels,wang2017dominant,kanezaki2018rotationnet,wei2020view} seek to find a more effective way to aggregate the view features. Specifically, Recurrent Cluster Pooling CNN (RCPCNN)~\citep{wang2017dominant} and Group-View CNN (GVCNN)~\citep{feng2018gvcnn} group views into multiple sets and conduct pooling within each set. SeqViews2SeqLabels~\citep{han2019seqview2seqlabels} and 3D2SeqViews~\citep{han20193d2seqvies} model the view order through recurrent neural network. Multi-Loop-View Convolutional Neural Network (MLVCNN)~\citep{jiang2019mlvcnn} designs a hierarchical view-loop-shape architecture and applies max pooling to the outputs from all hidden layers in the Long Short-Term Memory (LSTM) to obtain loop-level descriptors. View-based Graph Convolutional Network (View-GCN)~\citep{wei2020view} models the view-based relations through a recurrent neural network. Multi-view Harmonized Bilinear Network (MHBN)~\citep{yu2018multi} and Multi-View VLAD Network (MVLADN)~\citep{yu20213d} observe the limitation of view-based pooling, formulates the view-based 3D object recognition into a set-to-set matching problem, and investigates in patch-level pooling. Even though existing view-based methods have achieved excellent performance in 3D recognition,  they ignore the commutations between patches from different projected views when generating the patch/view features. As we know, a patch in a project view might be closely relevant to a patch in another view. A communication between relevant patches from different projected views might be beneficial to 3D object recognition.

Observing the limitations of existing methods, in this work, we propose round-roll MLP (R$^2$-MLP) for achieving communications between patches from different views. The key novelty of R$^2$-MLP architecture is the round-roll module. It shifts a part of features from each view to its adjacent views in a round-roll manner. Through the round-roll shifting, the patches in a specific projected view can exchange the information with the patches in its neighboring projected views, making the communications across the views feasible. Meanwhile, we adopt the spatial-shift operation in S$^2$-MLP architecture for communications between patches within each view.  In the implementation, we simultaneously achieve the round-roll operation for communications across views and the spatial-shift operation for communications within views in a single component, termed the round-roll (R$^2$) module.

\newpage

\begin{figure}[ht]
\centering
\includegraphics[width=.6\linewidth]{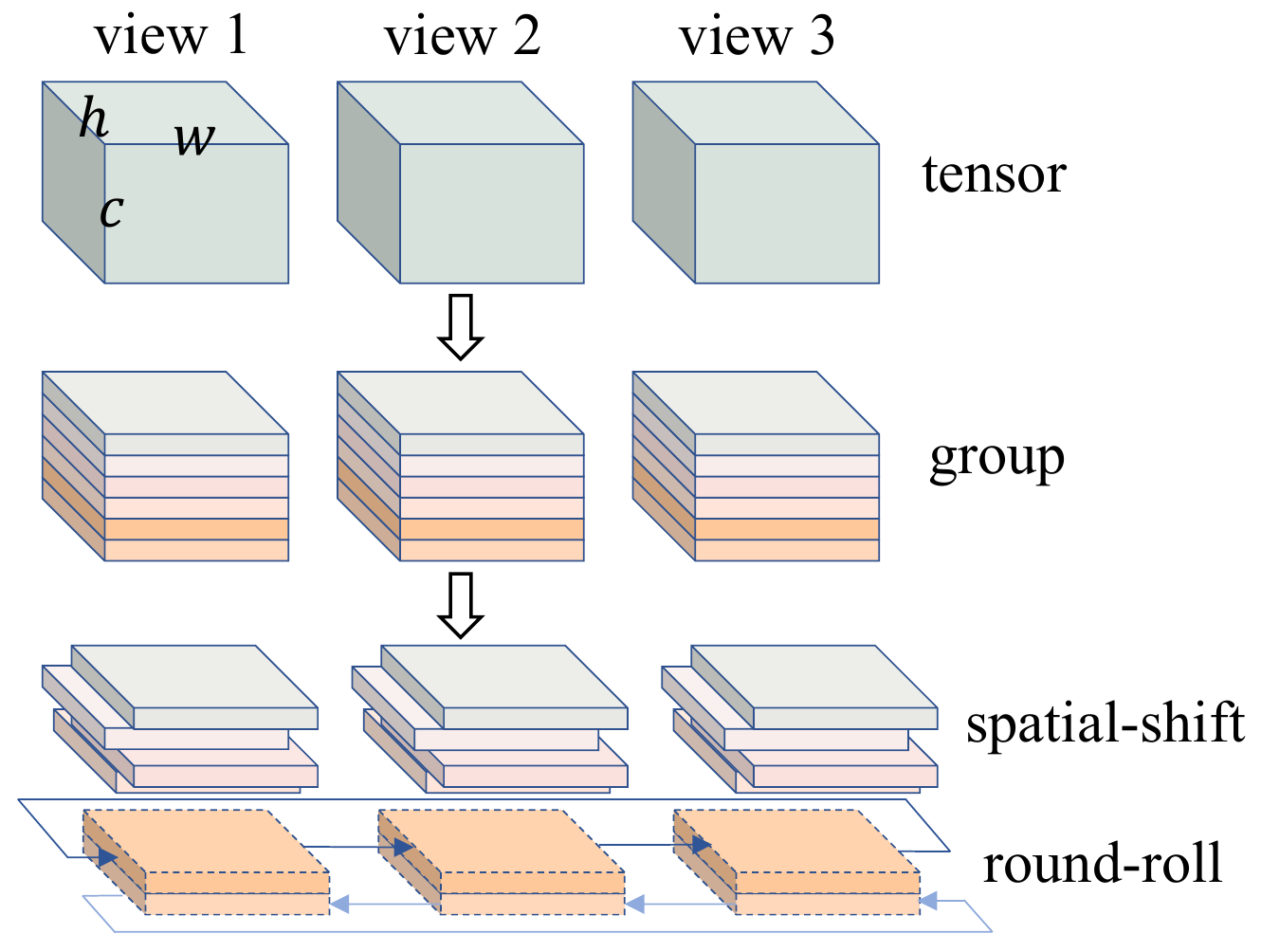}

\vspace{0.1in}

\caption{
The round-roll module. A 3D object is projected into $v$ views, generating a feature tensor $\bm{\mathcal{T}} \in \mathbb{R}^{v \times w \times h \times c}$, where $v, w, h, c$ denote the view number, width, height, and channel axis, separately. $c$ channels are equally divided into $6$ groups. The first four groups of channels are shifted along the width and height dimensions for communications between patches within each view. The last two groups are shifted along the view dimension for communications between patches from different views. The shifting operation along the view dimension is in a round-roll manner. Specifically, in the case visualized in the figure, the round-roll shift is in this manner: view 1 $\to$ view 2 $\to$ view 3 $\to$ view 1.
}\label{fig:roll}\vspace{0.2in}
\end{figure}

Figure~\ref{fig:roll}  illustrates the proposed round-roll module. Based on the R$^2$ module and MLPs,  we build our round-roll MLP architecture.
We propose a computation-free round-roll operation to interact with patch features from different views.
Compared to convolution and self-attention, shifting and rolling have benefits on computation. The complexity of parameters and FLOPs are shown in Table~\ref{tab:complexity}.

\vspace{0.4in}

\begin{table}[h!]
\caption{
$V$, $H$, $W$, and $C$ are the feature map's number of views, height, width, and channel numbers. $K$ is the convolutional kernel size.
R$^2$-MLP is less computationally intensive than the self-attention layer due to the lack of calculations such as the attention matrix. As for the parameter cardinality, we propose computation-free shifting and rolling operations to interact with the patch features between views.
R$^2$-MLP has more advantages in parameters and FLOPs than convolution and self-attention.
}
\label{tab:complexity}
\centering
\begin{tabular}{lll}
\toprule
Operation & Params & FLOPs \\
\midrule
Convolution & $\mathcal{O}(V K^2 C^2)$ & $\mathcal{O}(V K^2 H W C^2)$ \\
Self-Attention & $\mathcal{O}(3 V C^2)$ & $\mathcal{O}(V H^2 W^2 C^2)$ \\
Round-Roll MLP & $\mathcal{O}(V C^2)$ & $\mathcal{O}(V H W C^2)$\\
\bottomrule
\end{tabular}
\end{table}

\newpage

In summary, our contributions are three-fold:
\begin{itemize}
    \item We introduce the round-roll module. It achieves communications between adjacent views by exchanging a part of features of patches in adjacent views.
     \item We introduce a conceptually simple MLP-based architecture for 3D object recognition. As far as we know, this is the first work to utilize MLP-based models in this field.
    \item We have conducted experiments on the popular ModelNet40 and ModelNet10 datasets. Our R$^2$-MLP achieves the compelling results on ModelNet40  and outperforms the state-of-the-art methods on ModelNet10.
\end{itemize}

\section{Related Works}
\subsection{3D object recognition}
Existing mainstream 3D object recognition methods can be categorized into three groups according to the input mode:
volume-based methods~\citep{wu20153d,qi2016volumetric},
point-based methods~\citep{qi2017pointnet,qi2017pointnet++}
and view-based methods~\citep{wang2017dominant,feng2018gvcnn,kanezaki2018rotationnet,han2019seqview2seqlabels,han20193d2seqvies,wei2020view}.
Among them, volume-based methods quantize the 3D object into regular voxels and conduct  3D convolutions on voxels. Nevertheless, it is computationally expensive for 3D convolution when the resolution is high. To improve  efficiency,  volume-based methods  normally conduct low-resolution quantization. Therefore, there is inevitable information loss. At the same time, point-based methods directly model the cloud of points, efficiently achieving competitive performance. Although the volume-based and point-based methods utilize the spatial information of 3D objects, their practical applications are limited by the computational cost. View-based methods do not rely on complex 3D features. They project a 3D object into multiple 2D views. They model each view through the backbone for 2D image understanding, to obtain view  or patch features. View-based methods have another advantage of utilizing the backbone model trained on the mass amount of image data. The input of our method is multiple views. Thus, we mainly review view-based methods in the following.

MVCNN~\citep{su2015multi} pioneers exploiting CNN for modelling multiple views. It aggregates the view features from CNN through max-pooling. As an improvement to MVCNN~\citep{su2015multi}, MVCNN-MultiRes~\citep{qi2016volumetric} exploits views projected from multi-resolution settings, boosting the recognition accuracy. Different from the methods above taking fixed viewpoints, Pairwise~\citep{johns2016pairwise} decomposes the sequence of projected views into several pairs and models the pairs through CNN. GIFT~\citep{bai2016gift} proposes a more comprehensive virtual camera setting. It represents each 3D object by a set of view features. It determines the similarity between two 3D objects by matching two sets of view features through a devised matching kernel. RotationNet~\citep{kanezaki2018rotationnet} allows sequential input of views and treats viewpoints as latent variables to boost the recognition performance. RCPCNN~\citep{wang2017dominant} groups views into multiple sets and concatenates the set features as the 3D object representation. GVCNN~\citep{feng2018gvcnn} also groups multiple projected views into multiple sets. But it adaptively assigns a higher weight to the group containing crucial visual content to suppress the noise. Instead of CNN, Seqviews2seqlabels~\citep{han2019seqview2seqlabels} and 3D2SeqViews~\citep{han20193d2seqvies} exploit the view order besides visual content through recurrent neural network. Later on, View-GCN~\citep{wei2020view} models the relations between views by a graph convolution network. MHBN~\citep{yu2018multi} and MVLADN~\citep{yu20213d} investigate pooling patch-level features to generate the 3D object recognition.  Relation Network~\citep{yang2019learning} enhances each patch feature by patches from all views through a reinforcement block plugged in the rear of the network.
Besides CNN, a more recent work MVT~\citep{chen2021mvt} introduces the Transformer block and achieves state-of-the-art.
Unlike the abovementioned methods, we introduce a conceptually simple MLP-based architecture for 3D object recognition.

\subsection{MLP-based architectures}
Recently, models with only MLPs and skip connections has been a new trend for visual recognition tasks~\citep{tolstikhin2021mlp,touvron2022resmlp,yu2022s2mlp,yu2021rethinking,yu2021s2mlpv2}. MLP-based models resort to neither convolution layers. As a replacement, they use MLP layers to aggregate the spatial context. MLP-mixer~\citep{tolstikhin2021mlp} is the first work for exploiting pure MLP-based vision backbone. It achieves the communications between patches through a token-mixing MLP built upon two fully-connected layers. Concurrently, Res-MLP~\citep{touvron2022resmlp} simplifies the token-mixing MLP to a single fully-connected layer and utilizes more layers to build a deeper architecture, achieving a higher recognition accuracy. To further improve the accuracy, Spatial-shift MLP backbone (S$^2$-MLP)~\citep{yu2022s2mlp} adopts the spatial-shift operation for cross-patch communications. AS-MLP~\citep{lian2022as} also conducts the spatial-shift operation but shifts the features along two axes separately. S$^2$-MLP v2~\citep{yu2021s2mlpv2}, an improved version of S$^2$-MLP, uses a pyramid structure to boost the image recognition performance. CCS-MLP~\citep{yu2021rethinking} rethinks the design of token-mixing MLP and proposes a circulant channel-specific MLP. Specifically, it devises the weight matrix of token-mixing MLP as a circulant matrix, taking fewer parameters. Meanwhile, CCS-MLP can efficiently compute the multiplication between vector and circulant matrix through Fast Fourier Transform (FFT). GFNet~\citep{rao2021global} adopts 2D FFT to map the feature map into the frequency domain and mix frequency-domain features through MLP.

Though the existing methods perform well on image recognition, they are not proposed for multi-view 3D object recognition. Therefore, we explore how to efficiently aggregate patch features from different views on the MLP-based architecture. We propose the round-roll module to shift and roll the tensor consisting of patch features. Then patches can communicate with patches from the same view and the patches from other views.

\section{Method}

\subsection{Spatial-shift operation}

As is proposed in~\citet{yu2022s2mlp}, the spatial-shift operation is spatial-agnostic and maintains a local receptive field.
Given an input tensor $\bm{X} \in \mathbb{R}^{w \times h \times c}$, where $w$ is width, $h$ is height and $c$ represents the channel. First $\bm{X}$ is split into four parts $\{\bm{X}_i\}_{i=1}^4$ along channel dimension. Then each part is shifted along one direction:
\begin{align}
\begin{split}
     \bm{X'}[1\!: ,\ : ,\ :\!c/4] &= \bm{X}[:\!w\!-\!1 ,\ : ,\ :\!c/4], \\
     \bm{X'}[:\!w\!-\!1 ,\ : ,\ c/4\!:\!c/2] &= \bm{X}[1\!: ,\ : ,\ c/4\!:\!c/2], \\
     \bm{X'}[: ,\ 1\!: ,\ c/2\!:\!3c/4] &= \bm{X}[: ,\ :\!h\!-\!1 ,\ c/2\!:\!3c/4], \\
     \bm{X'}[: ,\ :\!h\!-\!1 ,\ 3c/4\!:] &= \bm{X}[: ,\ 1\!: ,\ 3c/4\!:].
\end{split}
\end{align}
After spatially shifting, each patch absorbs the visual content from its connecting patches. It is parameter-free and efficient for computation by simply shifting channels from a patch to its adjoining patches.

\begin{figure}[ht]

\centering
\mbox{
     \includegraphics[width=5in]{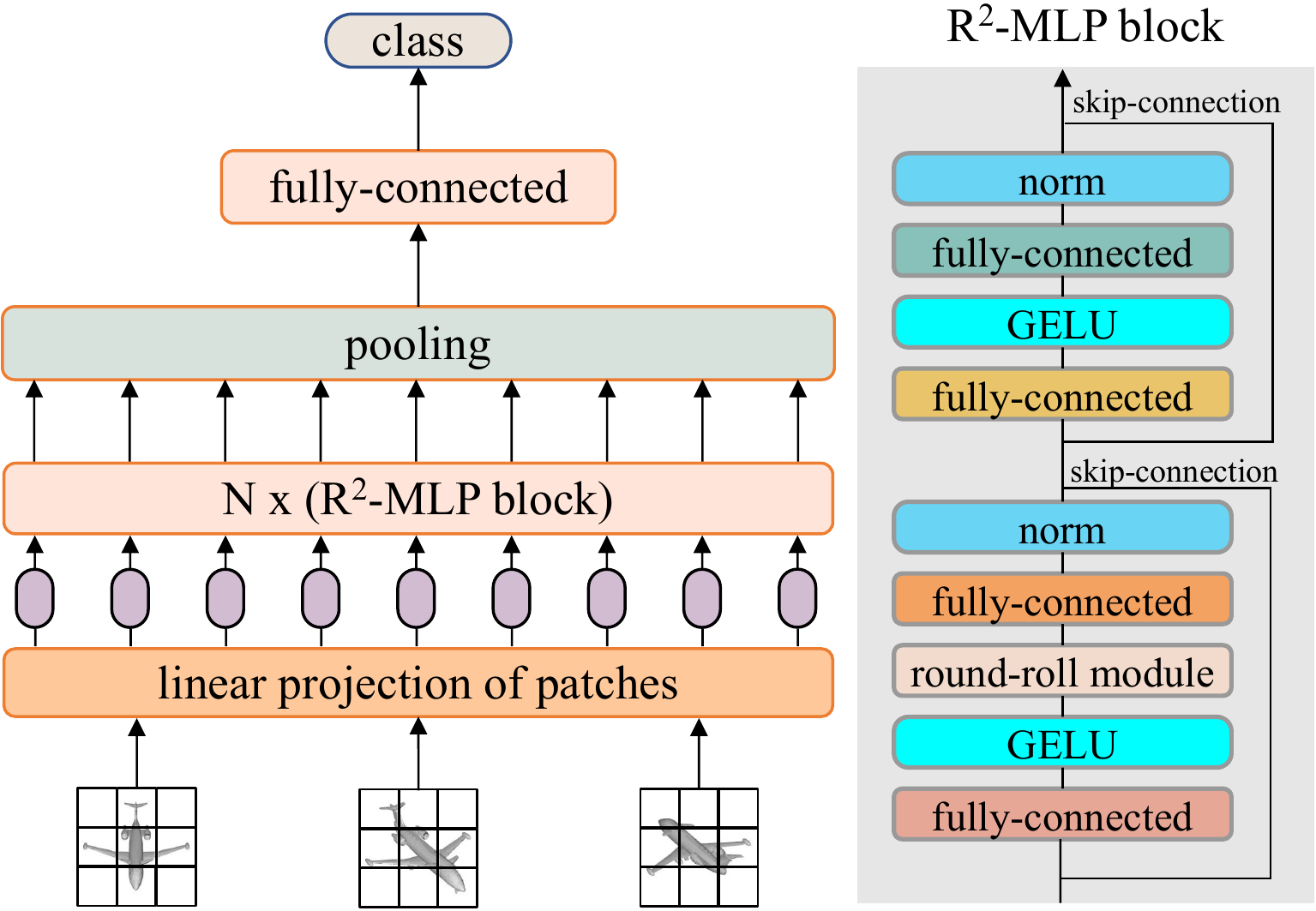}
}

\vspace{0.2in}

     \caption{
    The Round-Roll MLP (R$^2$-MLP) architecture. A 3D object is projected into multiple views. Each view is split into non-overlap patches. Then we project each patch into an embedding vector. The patch feature vectors are processed by a stack of $N$ R$^2$-MLP blocks for communication. Then these features are pooled into a global feature vector. We finally use the linear layer head for classification.}
     \label{fig:overview}
 \end{figure}

\subsection{Round-Roll MLP Architecture}
Figure \ref{fig:overview}  provides an overview of our method. It contains a patch embedding layer, a stack of $N$ R$^2$-MLP blocks, a pooling layer, and a fully-connected layer for classification.

\vspace{0.2in}\noindent\textbf{Patch embedding layer.}
For each 3D object, we project it into $v$ views. Each view image $V^j (j=1, \dots, v)$ is of $W \times H \times 3$ size. Identically to S$^2$-MLP, we split each view into $w \times h$ patches and project each patch of size $p \times p \times 3$ to a linear layer and obtain a $c$ dimensional vector:
\begin{equation}
    \bm{x}_i^j = \bm{W}_0 \bm{p}_i^j + \bm{b}_0, ~i = 1 \dots wh,~ j = 1 \dots v,
\end{equation}
where $\bm{W}_0 \in \mathbb{R}^{c \times 3p^2}$ is the weight matrix and $\bm{b}_0$ is the bias vector.
In total, each 3D object generates $vwh$ patches.

\vspace{0.2in}\noindent\textbf{R\texorpdfstring{$^2$}{\texttwosuperior}-MLP block.}
Our architecture stacks $N$  R$^2$-MLP blocks with the same structure. Each block contains four fully-connected layers, two layer normalization (LN) layers, two GELU activation layers, and the proposed round-roll module. Meanwhile, after each LN layer is paralleled with a skip-connection~\citep{he2016deep}. As we have introduced the details of LN and GELU in the previous section,  we only introduce the round-roll module here.

\begin{figure}[t!]
\centering
\includegraphics[width=6in]{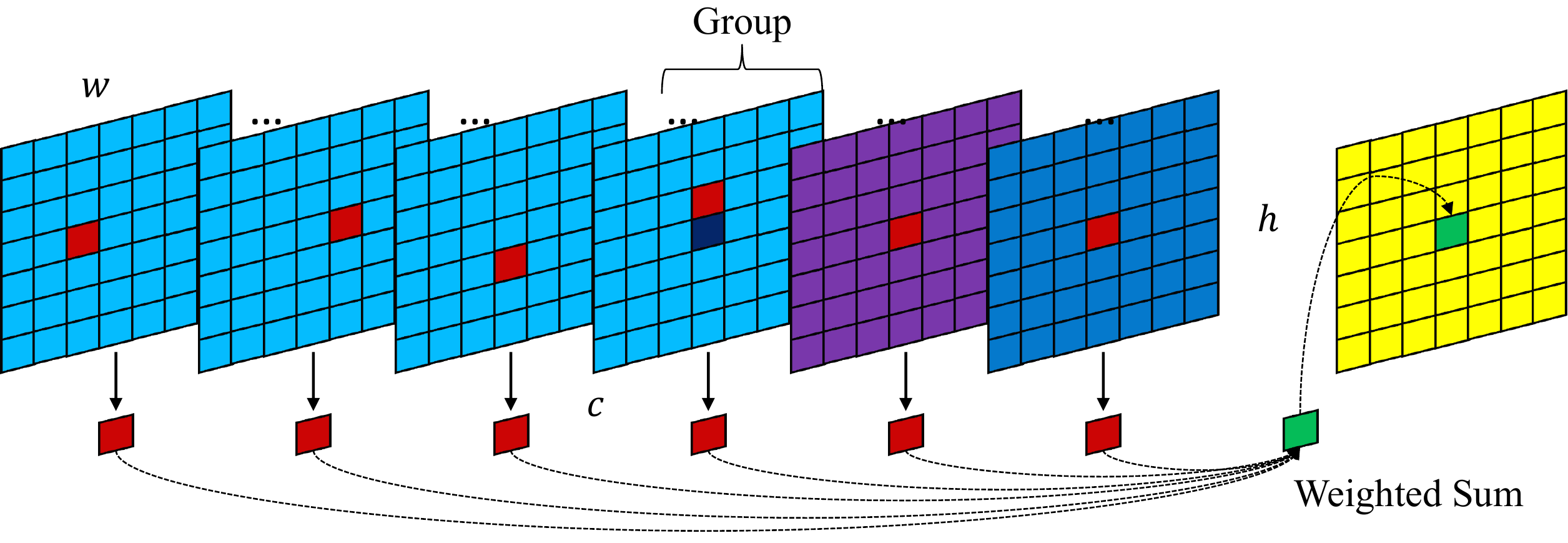}
\caption{
Illustrative round-roll operation of one fully-connected layer. The light blue, purple, and blue denote the current, previous, and next view feature channels separately. The yellow is the output features. The dark blue dot represents the position of interest, the dark red denotes other features used in the calculation process, and the green represents the corresponding output feature.``Group'' means that the channel is split into different groups.
}
\label{fig:illustrative}
\end{figure}

Similar to the spatial-shift module~\citep{yu2022s2mlp}, our round-roll module achieves cross-patch communications through shifting channels. Unlike the S$^2$ module, which only shifts the channels along the height and width dimension, our round-roll module also shifts channels along the view dimension for communications between patches across different views. Suppose we have an input tensor $\bm{\mathcal{T}} \in \mathbb{R}^{v \times w \times h \times c}$, where $v$ is the number of views for each 3D object model, $w$ denotes the weight, $h$ is the height, and $c$ is the number of channels.
Our proposed round-roll module groups $c$ into several groups, shifting and rolling different channel groups in different directions. The feature map, after the rolling, aligns different views' token features to the same channel, and then the interaction of the view information can be realized after channel projection.
The calculation is visualized in Figure~\ref{fig:illustrative}.
Specifically, the round-roll operation includes three steps: 1) uniformly split the channels into six groups, 2) shift the first four groups of channels on weight and height axes, and 3) roll the last two groups of channels on the view axis. To be specific, we equally split $\bm{\mathcal{T}}$ along channel dimension into six thinner tensors $\{ \bm{\mathcal{T}}_t \}_{t=1}^6$, where $\bm{\mathcal{T}}_t \in \mathbb{R}^{v \times w \times h \times c/6}$. For $\bm{\mathcal{T}}_1$, we shift it along the width axis by +1. Meanwhile, we shift the $\bm{\mathcal{T}}_2$ along the width axis by -1. $\bm{\mathcal{T}}_3$ and $\bm{\mathcal{T}}_4$ are shifted by +1 and -1 along the height axis separately. That is, the shift operations on $\{ \bm{\mathcal{T}}_t \}_{t=1}^4$ are similar to that in  the spatial-shift module~\citep{yu2022s2mlp}. While for $\bm{\mathcal{T}}_5$ and $\bm{\mathcal{T}}_6$, we adapt the round-roll operation on these two components along the view dimension. Specifically, we roll $\bm{\mathcal{T}}_5$ on the view axis by +1. In parallel, $\bm{\mathcal{T}}_6$ is rolled by -1 along the view axis.

\begin{align}
\label{rreq}
\begin{split}
     \bm{\mathcal{T'}}[:,1\!:,:,:\!c/6] &= \bm{\mathcal{T}}[:,:\!w\!-\!1,:,\ :\!c/6], \\
     \bm{\mathcal{T'}}[:,:\!w\!-\!1,:,c/6\!:\!c/3] &= \bm{\mathcal{T}}[:,1\!:,:,c/6\!:\!c/3], \\
     \bm{\mathcal{T'}}[:,:,\!1\!:,c/3\!:\!c/2] &= \bm{\mathcal{T}}[:,:,:\!h\!-\!1,\!c/3\!:\!c/2], \\
     \bm{\mathcal{T'}}[:,:,\:\!h\!-\!1,c/2\!:\!2c/3] &= \bm{\mathcal{T}}[:,:,1\!:,c/2\!:\!2c/3], \\
     \bm{\mathcal{T'}}[1\!:,:,:,2c/3\!:\!5c/6] &= \bm{\mathcal{T}}[:\!v\!-\!1,:,:,\!2c/3\!:\!5c/6], \\
     \bm{\mathcal{T'}}[0,:,:,2c/3\!:\!5c/6] &= \bm{\mathcal{T}}[v\!-\!1,:,:,\!2c/3\!:\!5c/6], \\
     \bm{\mathcal{T'}}[:\!v\!-\!1,:,\:\!h\!-\!1,5c/6\!:] &= \bm{\mathcal{T}}[1\!:,,:,1\!:,5c/6\!:],\\
     \bm{\mathcal{T'}}[v\!-\!1,:,:,5c/6\!:] &= \bm{\mathcal{T}}[0,:,:,5c/6\!:].\\
\end{split}
\end{align}

\section{Experiments}

\subsection{Settings}
\begin{figure}[htb!]
\vspace{0.2in}

    \centering
    \begin{subfigure}[b]{0.45\textwidth}
        \centering
        \includegraphics[width=.7\textwidth]{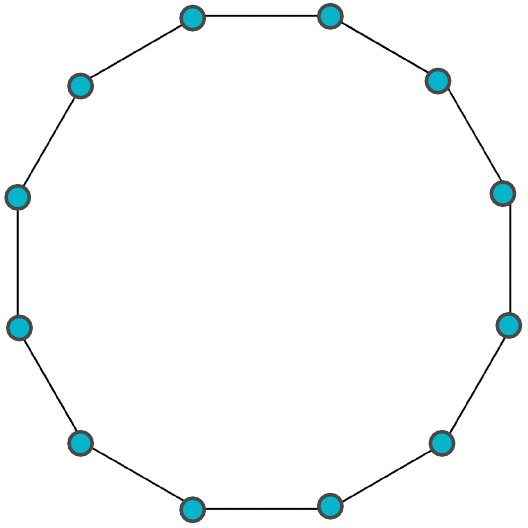}
        \caption{12 view setting.}
        \label{fig:12-view}
    \end{subfigure}
    \hfill
    \begin{subfigure}[b]{0.45\textwidth}
        \centering
        \includegraphics[width=.75\textwidth]{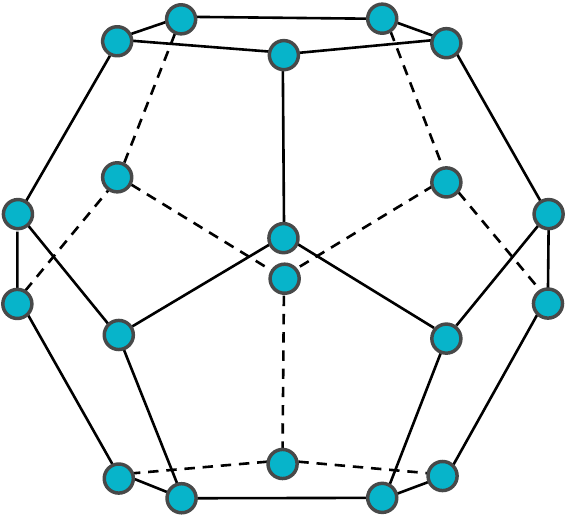}
        \caption{20 view setting.}
        \label{fig:20-view}
    \end{subfigure}
\caption{Graph structure of 12 view and 20 view settings.
}
\label{fig:graph}
\end{figure}

\noindent\textbf{Datasets.}
We conduct the experiments on the popular benchmark ModelNet40~\citep{chang2015shapenet} and ModelNet10~\citep{chang2015shapenet} datasets. ModelNet40 has 40 categories totaling 12311 3D models, divided into 9843 training data and 2468 test data. ModelNet10 is a subset of ModelNet40, including 10 categories. It has 3991 training data and 908 test data. We have two viewpoints settings: 12 views and 20 views. For 12 view setting, we render the 3D mesh models by placing 12 centroid pointing virtual cameras around the mesh every 30 degrees with an elevation of 30 degrees from the ground plane, as shown in Figure~\ref{fig:12-view}. We utilize the rendered view images provided by~\citet{wang2017dominant}. This viewpoint setting is adopted when we feed $\leq 12$ for each 3D model. 20 view setting places 20 virtual cameras on vertices of a dodecahedron encompassing the object, as shown in Figure~\ref{fig:20-view}. In experiments, we use the rendered images offered by~\citet{kanezaki2018rotationnet}.

\vspace{0.2in}\noindent\textbf{Evaluation metric.} We use instance accuracy, the ratio of the number of correctly classified objects to the total number of objects, as the metric of the classification task.

\vspace{0.2in}\noindent\textbf{Implementation details.} We utilize the training strategy offered by DeiT~\citep{touvron2021training}. Specifically, we set the batch size as 64. We train our network with the AdamW optimizer with a learning rate of 0.003 and weight decay of 0.05. The learning rate is scheduled with a linear warmup and cosine decay for 1000 epochs. We also use data augmentation methods, viz., label smoothing, Auto-Augment, Rand-Augment, random erase, mixup and cutmix. We also adopt Stochastic depth. The customization is set identically as DeiT~\citep{touvron2021training}.

\subsection{Ablation studies}
Unless specified, we use 6 views for each 3D model. We stack 36 layers of round-roll modules, namely R$^2$-MLP-36, and train it from scratch. We perform ablation studies on the ModelNet10 dataset.

\begin{figure*}[t!]
\centering
\includegraphics[width=6.5in]{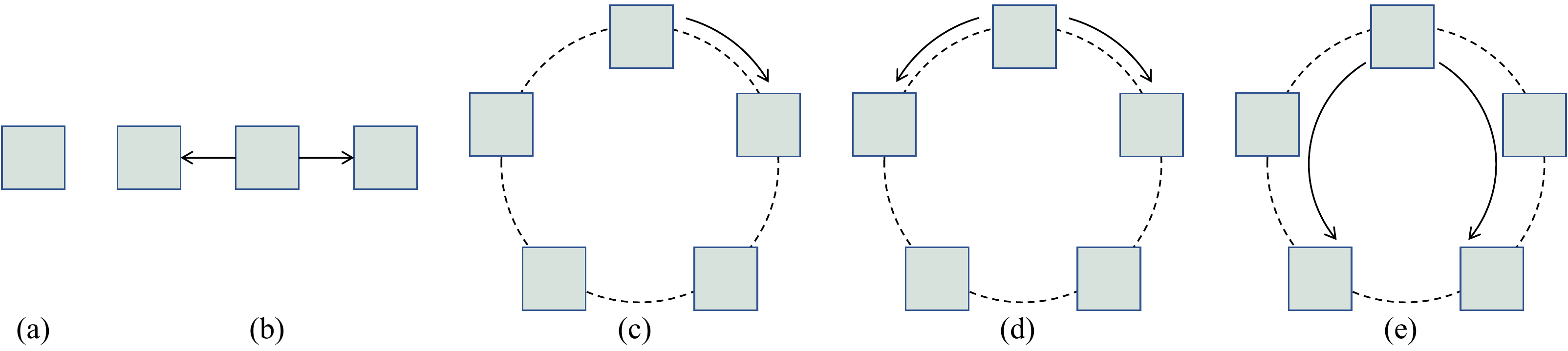}

\begin{tabular}{lcccccc}
\toprule
setting & (a) & (b) & (c) & (d) & (e) \\
\midrule
accuracy  & 92.62 & 92.84 & 93.94 & \textbf{94.71} & 94.49 \\
\bottomrule
\end{tabular}

\vspace{0.1in}

\caption{The different settings to ablate the round-roll module. (a) spatial-shift only and no interrelation between views; (b) shift 1 stride both forward and backward; (c) roll 1 stride forward; (d) roll 1 stride both forward and backward; (e) roll 2 strides both forward and backward. We observe that the best performance is (d) when rolling feature tensor along view dimension 1 stride forward and backward.} \label{tab:roll}\vspace{0.1in}
\end{figure*}

\vspace{0.2in}\noindent\textbf{Ablation on the round-roll module.} We first ablate our proposed round-roll module. We illustrate different settings in Figure~\ref{tab:roll}. In setting (a), we adopt spatial-shift operation only, which is identical to~\citet{yu2022s2mlp} and no interrelation occurs between views. We treat setting (a) as a baseline. We shift the tensor 1 stride both forward and backward in (b). The accuracy increases from 92.62\% (a) to 92.84\% (b). From (c) to (e) we roll the tensor instead of shift. Rolling operation makes it available for the first view and last view to interrelate with each other. And we can observe a significant improvement when adopting rolling operation. All (c), (d), and (e) settings perform better than (b). Rolling both forward and backward, (d) and (e), reach a noticeable improvement. But rolling 1 stride on two directions has higher accuracy than 2 strides. We think this might be because, with less stride, the view could interrelate with closer adjacent views that are more similar to themselves. So, stride 1 in (d) could perform better than 2 in (e). And we use (d), rolling 1 stride on two directions, as the default.

The round-roll module splits the feature tensor into 6 groups along channel axis $c$. The first four groups are shifted along the width and height dimensions while the last two groups are shifted along the view dimension.

\begin{table*}[h]

\vspace{0.2in}

\caption{
Ablation on R$^2$-MLP block's group splitting. We assign four groups for spatial-shift and two groups for round-roll.
The channel $c$ is 384 in total.
The first row is for each spatial-shift group, and the second is for each round-roll group. Equally divided is the best.
}
\label{tab:group}
\centering
\begin{tabular}{lccccccc}
\toprule
 spatial-shift & 0 & 16 & 32 & 48 & 64 & 80 & 96 \\
 round-roll & 192 & 160 & 128 & 96 & 64 & 32 & 0 \\
 \midrule
 accuracy & 89.21 & 93.29 & 93.78 & 94.43 & \textbf{94.71} & 94.49 & 92.62 \\
\bottomrule
\end{tabular}\vspace{0.2in}
\end{table*}

We ablate the number of channels assigned to each group in Table~\ref{tab:group}. The results show that equally dividing the channels guarantees the best performance, and we set it as the default.

\newpage

\begin{table}[h]
\caption{The influence of the number of views on the ModelNet10 dataset with R$^2$-MLP-36 model trained from scratch.}
\label{tab:view}
\centering
\begin{tabular}{cccc}
\toprule
views & 1 & 3 & 6 \\
\midrule
acc. & 90.42 & 94.05 & \textbf{94.71} \\
\bottomrule
\end{tabular}
\end{table}

\vspace{0.2in}\noindent\textbf{The influence of the number of views.} Table~\ref{tab:view} ablates the effect of the number of views on the average instance accuracy of our method on the ModelNet10 dataset. More projected views lead to better classification accuracy. Specifically, using a single view, R$^2$-MLP-36 model without pre-training only achieves a $90.42\%$ recognition accuracy, whereas using 6 views achieves a $94.71\%$ recognition accuracy. Better performance is expected since more views provide more visual information for a 3D object and benefit recognition.

\vspace{0.2in}\noindent\textbf{Ablating the depth.}
We ablate the influence of the network depth on the computation cost and recognition accuracy. Following~\citep{yu2022s2mlp}, we testify three models of variant depth: R$^2$-MLP-12, R$^2$-MLP-24, and R$^2$-MLP-36.
The models are trained from scratch with 6 views on the ModleNet10 dataset. We also count the number of parameters, FLOPs, and inference throughput of each model. We set the batch size (the number of 3D objects per batch) as 4. We use an NVIDIA TITAN X Pascal GPU during the inference. The results are shown in Table~\ref{tab:arch}. We could see that the deeper model takes more parameters and FLOPs and achieves higher accuracy. By default, we adopt the $36$-layer.

\begin{table}[h]
\caption{The influence of different architecture. The inference throughput is measured as the number of 3D objects processed per second on an NVIDIA TITAN X Pascal GPU.}\label{tab:arch}
\centering
\begin{tabular}{llll}
\toprule
Model & R$^2$-MLP-12 & R$^2$-MLP-24 & R$^2$-MLP-36 \\
\midrule
layers & 12 & 24 & 36 \\
params & 18.0M & 35.8M & 53.5M \\
FLOPs & 21B & 42B & 63B \\
throughput & 69 & 41 & 27 \\
accuracy & 94.38 & 94.49 & \textbf{94.71} \\
\bottomrule
\end{tabular}
\end{table}

\vspace{0.1in}\noindent\textbf{The influence of pre-training.}
A straightforward approach to boost the performance of our R$^2$-MLP is to initialize its weights by the model pre-trained on a large-scale image recognition dataset. To this end, we use the weights of S$^2$-MLP-deep~\citep{yu2022s2mlp} to initialize our R$^2$-MLP.

\begin{table}[htb!]
\caption{Influence of ImageNet pre-training.}
\label{tab:pretrain}\centering
\begin{tabular}{ccc}
\toprule
R$^2$-MLP-36 & w/o pre-train & w/ pre-train \\
\midrule
accuracy & 94.71 & \textbf{95.93} \\
\bottomrule
\end{tabular}
\end{table}

Table~\ref{tab:pretrain} shows the results on the ModelNet10 dataset.  When fine-tuning the pre-trained model, the accuracy is 95.93\% compared to 94.71\% when training from scratch. Higher accuracy shows that the knowledge from ImageNet benefits our R$^2$-MLP on the 3D object recognition task.

\vspace{0.2in}\noindent\textbf{Hybrid of S\texorpdfstring{$^2$}{\texttwosuperior}-MLP block and R\texorpdfstring{$^2$}{\texttwosuperior}-MLP block.}
To bridge the architecture gap between the pre-trained S$^2$-MLP model and our R$^2$-MLP model for a better transfer learning performance, a plausibly more reasonable manner is to use  S$^2$-MLP blocks in the shallow layers and R$^2$-MLP blocks in the deep layers. Specifically, we stack several S$^2$-MLP blocks and follow R$^2$-MLP blocks. The results are shown in Table~\ref{tab:hybrid}.

\begin{table}[ht]
\caption{Influence of R$^2$-MLP depth.}\label{tab:hybrid}
\centering
\begin{tabular}{ccccc}
\toprule
S$^2$-MLP blocks & 24 & 12 & 0 \\
R$^2$-MLP blocks & 12 & 24 & 36 \\
\midrule
accuracy & 94.99 & 95.48 & \textbf{95.93} \\
\bottomrule
\end{tabular}
\end{table}

We keep the total number of blocks as $36$ and gradually increase the R$^2$-MLP blocks. While we have more R$^2$-MLP blocks, the accuracy increases accordingly. The R$^2$-MLP only architecture achieves higher performance than the hybrid architecture. The efforts to bridge the architecture gap from the hybrid architecture cannot achieve satisfactory results. So we do not adopt this hybrid architecture by default.

\vspace{0.2in}\noindent\textbf{Influence of fusion ways on patch features.}
After a stack of $N$ R$^2$-MLP blocks, we obtain a tensor $\bm{\mathcal{T}} \in \mathbb{R}^{v\times w \times h \times c}$, which contains $vwh$ $c$-dimension patch features. We need to pool these patch features in the tensor into a $c$-dimension feature for the final recognition. For the easiness of the illustration, we term the pooling on the patches along the height dimension and the width dimension as spatial pooling and term the pooling on patches along the view dimension as view pooling.

Considering the order and pooling way, we investigate six different combinations on the spatial level and the view level: (i) mean view pooling followed by max spatial pooling; (ii)  max view pooling followed by mean spatial pooling; (iii) max view pooling followed by max spatial pooling; (iv) mean view pooling followed by max spatial pooling; (v) max spatial pooling followed by mean spatial pooling; (vi) mean spatial pooling followed by max view pooling.

\begin{table}[ht]
\caption{Influence of fusion ways on patch features.}
\label{tab:pool}
\centering
\begin{tabular}{l@{}lc}
\toprule
\multicolumn{2}{c}{fusion way} & accuracy \\
\midrule
(i) & view mean $\rightarrow$ spatial mean & 94.11 \\
(ii) & view mean $\rightarrow$ spatial max & 94.49 \\
(iii) & view max $\rightarrow$ spatial max & 94.60 \\
(iv) & view max $\rightarrow$ spatial mean & \textbf{94.71} \\
(v) & spatial max $\rightarrow$ view mean & \textbf{94.71} \\
(vi) & spatial mean $\rightarrow$ view max & 94.16 \\
\bottomrule
\end{tabular}
\end{table}

Table~\ref{tab:pool} shows that using only mean pooling or maxing pooling in both spatial pooling and view pooling can not achieve competitive results. In contrast, the hybrid version (iv) and (v) achieve the highest accuracy. We set (iv) as the default.

\begin{table*}[t!]
\caption{Comparison with the present state-of-the-art methods on the ModelNet40 dataset. The best accuracy is bold.}\label{tab:sota}
\centering
\begin{tabular}{ccccc}
\toprule
Method & Views & ModelNet40 & ModelNet10 \\
\midrule
MVCNN~\citep{su2015multi} & 80 & 90.1  & - \\
RotationNet~\citep{kanezaki2018rotationnet} & 12 & 91.0 & 94.0 \\
GVCNN~\citep{feng2018gvcnn} & 12 & 93.1 & - \\
Relation Network~\citep{yang2019learning} & 12 & 94.3 & 95.3 \\
3D2SeqViews~\citep{han20193d2seqvies} & 12 & 93.4 & 94.7 \\
SeqViews2SeqLabels~\citep{han2019seqview2seqlabels} & 12 & 93.4 & 94.8 \\
MLVCNN~\citep{jiang2019mlvcnn} & 12 & 94.2 & - \\
CAR-Net~\citep{xu2021multi} & 12 & \textbf{95.2} & 95.8 \\
MVT-small~\citep{chen2021mvt} & 12 & 94.4 & 95.3 \\
\rowcolor{mygray} {\textbf{R$^2$-MLP-36 (Ours)}} & 6 & 94.7 & 95.9 \\
\rowcolor{mygray} {\textbf{R$^2$-MLP-36 (Ours)}} & 12 & 95.0  & \textbf{97.4} \\
RotationNet~\citep{kanezaki2018rotationnet}& 20 & 97.4 & 98.5 \\
View-GCN~\citep{wei2020view} & 20 & 97.6 & - \\
CAR-Net~\citep{xu2021multi} & 20 & \textbf{97.7} & 99.0 \\
MVT-small~\citep{chen2021mvt} & 20 & 97.5 & 99.3 \\
\rowcolor{mygray} {\textbf{R$^2$-MLP-36 (Ours)}} & 20 & \textbf{97.7} & \textbf{99.6} \\
\bottomrule
\end{tabular}
\end{table*}

\newpage

\subsection{Comparison with state-of-the-art methods}

In Table~\ref{tab:sota}, we compare our method with view-based methods including MVCNN~\citep{su2015multi}, RotationNet~\citep{kanezaki2018rotationnet}, GVCNN~\citep{feng2018gvcnn}, Relation Network~\citep{yang2019learning}, 3D2SeqViews~\citep{han20193d2seqvies}, SeqViews2SeqLabels~\citep{han2019seqview2seqlabels}, MLVCNN~\citep{jiang2019mlvcnn} and CAR-Net~\citep{xu2021multi}. With 20 views, our R$^2$-MLP model achieves the best performance compared with these methods. Specifically, on the ModelNet10 dataset, we achieve the highest recognition accuracy of 99.6\%. Furthermore, on the ModelNet40 dataset, we are the best (97.7\%) like CAR-Net~\citep{xu2021multi}.
With 12 views, we are the best on the ModelNet10 dataset and the second-best on the ModelNet40 dataset.
With 6 views we also achieve comparable results.
It is worth noting that the architecture of our R$^2$-MLP has advantages in parameters and computation FLOPS than the compared methods using CNNs, \eg CAR\-Net~\citep{xu2021multi}. Meanwhile, our MLP-based architecture takes fewer parameters and less inference time. We compare the R$^2$-MLP-36 inference time with its CNN counterparts MVCNN~\citep{su2015multi} and CAR-Net~\citep{xu2021multi}.

\begin{table}[ht]
\caption{Comparison of inference time on ModelNet40.
Our model's inference speed is faster by a large margin.}\label{tab:size}
\centering
\begin{tabular}{lcc}
\toprule
{Model} & CAR-Net \citep{xu2021multi} & R$^2$-MLP-36 (Ours) \\
\midrule
inference & 0.20 s & 0.07 s \\
\bottomrule
\end{tabular}\vspace{0.2in}
\end{table}

Additionally, in Table~\ref{tab:size} we measure the time for processing a 3D object of ModelNet40 on one NVIDIA V100 GPU, following the same setting adopted in \citep{xu2021multi}. The CAR-Net~\citep{xu2021multi} is 0.20 s while our R$^2$-MLP-36 is 0.07 s.

\subsection{3D shape retrieval}
Besides the classification task, we also evaluate our R$^2$-MLP on the well-known 3D shape retrieval benchmark: ModelNet40~\citep{chang2015shapenet}. The retrieval task in ModelNet40 is a general 3D shape retrieval task (within-domain task), in which the given query objects and target objects are all 3D models. We adopt the evaluation metric mean Average Precision (mAP) to measure the retrieval performance.
For this task, we follow the retrieval setup of MVCNN~\citep{su2015multi}. In particular, we consider the deep feature representation of the last layer before the classification head. We first use $L^2$-normalization on feature vectors.
We fine-tune R$^2$-MLP for classification; thus, retrieval performance is not optimized directly.
Instead of training it with a different loss suitable for retrieval, we use Local Fisher Discriminant Analysis for Supervised Dimensionality Reduction (LFDA)~\citep{sugiyama2007dimensionality} to project the feature into a more expressive space.
The reduced features describe a shape. During the inference, shape representations are used to retrieve the most similar shapes in the test set according to the Euclidean distance between pairwise shape-level representative features. \\

\begin{table}[htb!]
\caption{3D shape retrieval comparison with the state-of-the-art methods on the ModelNet40 dataset. R$^2$-MLP achieves the best retrieval performance among recent state-of-the-art methods.}\label{tab:retrieval}
\centering
\begin{tabular}{lcc}
\toprule
Method & Views & mAP \\
\midrule
LFD~\citep{chen2003visual} & Volume & 40.0 \\
3DShapeNets~\citep{chang2015shapenet} & Volume & 49.2 \\
PVNet~\citep{you2018pvnet} & Point & 89.5 \\
MVCNN~\citep{su2015multi} & 12 & 80.2 \\
MLVCNN~\citep{jiang2019mlvcnn} & 24 & 92.2 \\
MVTN~\citep{hamdi2021mvtn} & 12 & 92.9 \\
\rowcolor{mygray} {\textbf{R$^2$-MLP (Ours)}} & 12 & \textbf{93.0} \\
\rowcolor{mygray} {\textbf{R$^2$-MLP (Ours)}} & 20 & \textbf{93.1} \\
\bottomrule
\end{tabular}\vspace{0.2in}
\end{table}

\begin{figure}[htb!]
\centering
\includegraphics[width=.8\linewidth]{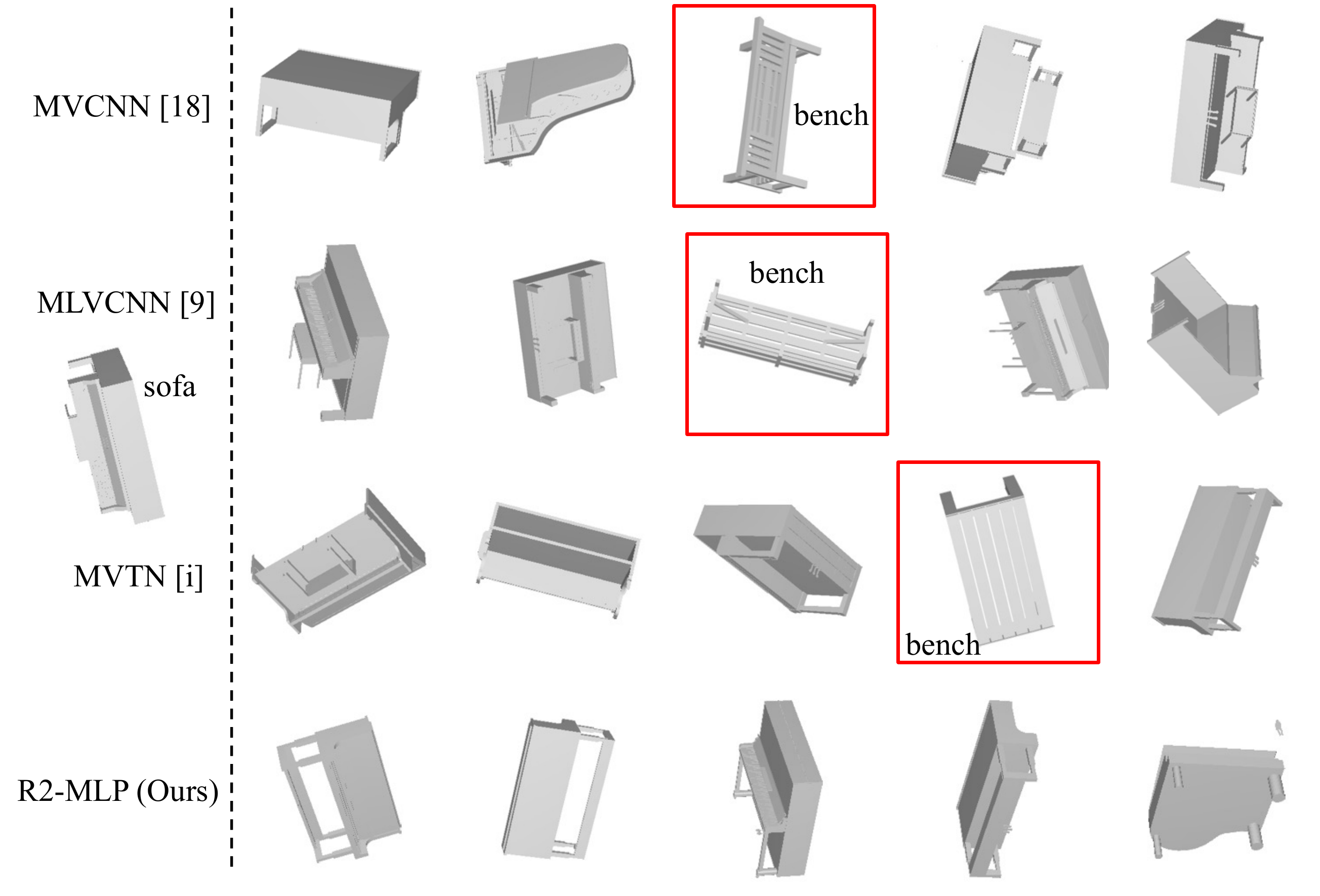}
\caption{
Qualitative examples for 3D object retrieval on ModelNet40: (left): we show the query object (sofa) from the test. (right): top five retrieved objects by different methods.
The negatively retrieved results are framed in red.
}
\label{fig:retr}\vspace{0.5in}
\end{figure}

Table~\ref{tab:retrieval} displays the comparison to the state-of-the-art methods. Volume-based LFD~\citep{chen2003visual} and 3DShapeNets~\citep{chang2015shapenet} have poor performance. In general, view-based methods perform much better than volume-based methods. Also, the point-based method PVNet~\citep{you2018pvnet} achieves a competitive result. Our method outperforms the  MLVCNN~\citep{jiang2019mlvcnn}, increasing from 92.2\% to 93.1\%, with fewer views. The outperformance shows our method's generality on the 3D shape retrieval task.
The qualitative examples of the retrieved results are shown in Figure~\ref{fig:retr}.

\section{Conclusion}
This paper proposes a Round-Roll MLP (R$^2$-MLP) for effective 3D object recognition. Considering the efficiency, we propose the R$^2$-MLP module, whose key component is shifting the feature tensor along the spatial dimension and rolling along the view dimension. The R$^2$-MLP module empowers each view to communicate with connected adjacent views, overcoming the limitations of existing S$^2$-MLP architectures with only spatial-shift operations on patches from the same view. Besides, the R$^2$-MLP module does not need extra computation cost and is efficient to deploy. Our proposed R$^2$-MLP architecture achieves SOTA on ModelNet40 and ModelNet10 datasets.

\bibliographystyle{plainnat}
\bibliography{refs_scholar}

\end{document}